**Evaluating Large Language Models Against Human Annotators in Latent Content Analysis: Sentiment, Political Leaning, Emotional Intensity, and Sarcasm**

Ljubiša Bojić[1]*, Ph. D.
Senior Research Fellow
Institute for Artificial Intelligence Research and Development of Serbia; University of Belgrade, Institute for Philosophy and Social Theory, Digital Society Lab

Olga Zagovora[2], Ph. D.
Senior Research Fellow
Rheinland-Pfälzische Technische Universität Kaiserslautern-Landau (RPTU), Landau; German Research Centre for Artificial Intelligence (DFKI)

Asta Zelenkauskaite[3]
Senior Researcher; Full Professor
Vilnius Gediminas Technical University, Vilnius; Drexel University

Vuk Vuković[4], Ph. D.
Associate Professor
University of Montenegro, Faculty of Dramatic Arts

Milan Čabarkapa[5], Ph. D.
Assistant Professor
University of Kragujevac, Faculty of Engineering

Selma Veseljević Jerković[6]

---

[1] Corresponding author
Email address: ljubisa.bojic@ivi.ac.rs; ljubisa.bojic@ifdt.bg.ac.rs
Address of correspondence: 1 Fruskogorska, Novi Sad, Serbia; Kraljice Natalije 45, 11000 Belgrade, Serbia
ORCID: 0000-0002-5371-7975

[2] Email address: olga.zagovora@rptu.de
Address of correspondence: Fortstraße 7, 76829 Landau; Trippstadter Str. 122, 67663 Kaiserslautern
ORCID: 0000-0002-4693-9668

[3] Email address: az358@drexel.edu
Address of correspondence: Saulėtekio al. 11, Vilnius, 10223 Vilniaus m. sav., Lithuania; 3201 Arch, 165 Philadelphia, PA 19104, USA
ORCID: 0000-0001-5762-4605

[4] Email address: vuk.vukovic@ucg.ac.me
Address of correspondence: Bajova 6, 81250 Cetinje, Montenegro
ORCID: 0000-0002-0766-1077

[5] Email address: mcabarkapa@kg.ac.rs
Address of correspondence: Sestre Janjić 6, 34000 Kragujevac, Serbia
ORCID: 0000-0002-2094-9649

[6] Email address: selma.veseljevic@untz.ba

Associate Professor
University of Tuzla, Faculty of Humanities and Social Sciences, Department of English Language and Literature

Ana Jovančević[7], Ph. D.
PostDoc
University of Limerick, Faculty of Education and Health Sciences, Department of Psychology

**Abstract**

In the era of rapid digital communication, vast amounts of textual data are generated daily, demanding efficient methods for latent content analysis to extract meaningful insights. Large Language Models (LLMs) offer potential for automating this process, yet comprehensive assessments comparing their performance to human annotators across multiple dimensions are lacking. This study evaluates the reliability, consistency, and quality of seven state-of-the-art LLMs, including variants of OpenAI's GPT-4, Gemini, Llama, and Mixtral, relative to human annotators in analyzing sentiment, political leaning, emotional intensity, and sarcasm detection. A total of 33 human annotators and eight LLM variants assessed 100 curated textual items, generating 3,300 human and 19,200 LLM annotations, with LLMs evaluated across three time points to examine temporal consistency. Inter-rater reliability was measured using Krippendorff's alpha, and intra-class correlation coefficients assessed consistency over time. The results reveal that both humans and LLMs exhibit high reliability in sentiment analysis and political leaning assessments, with LLMs demonstrating higher internal consistency than humans. In emotional intensity, LLMs displayed higher agreement compared to humans, though humans rated emotional intensity significantly higher. Both groups struggled with sarcasm detection, evidenced by low agreement. LLMs showed excellent temporal consistency across all dimensions, indicating stable performance over time. This research concludes that LLMs, especially GPT-4, can effectively replicate human analysis in sentiment and political leaning, although human expertise remains essential for emotional intensity interpretation. The findings demonstrate the potential of LLMs for consistent and high-quality performance in certain areas of latent content analysis.

## Introduction

In an era characterized by rapid digitization and the proliferation of online communication platforms, vast amounts of textual data are generated daily. This surge presents both an opportunity

Address of correspondence: Dr Tihomila Markovića, Tuzla, Bosnia and Herzegovina
ORCID: 0000-0003-3794-3961

[7] Email address: ana.jovancevic@ul.ie
Address of correspondence: National Technological Park Limerick V94 T9PX, Limerick, Ireland
ORCID: 0000-0002-1693-8891

and a challenge: while there is unprecedented access to public opinion and discourse, analyzing these data to extract meaningful insights requires substantial effort and resources. Latent content analysis, which involves decoding the underlying meanings, sentiments, and nuances in text, is crucial for understanding social dynamics, informing policy decisions, and guiding business strategies (Neuendorf, 2017). Automating this process could significantly enhance our ability to respond to societal needs promptly and effectively.

The societal implications of effectively analyzing textual content are profound. Sentiment analysis can reveal public opinion on policies or products, influencing governmental decisions and corporate strategies (Liu, 2012). Understanding political leanings aids in assessing electoral landscapes and fostering democratic engagement (DiMaggio, Evans, & Bryson, 1996). Detecting emotional intensity and sarcasm in communication is vital for mental health monitoring, customer service, and even national security (Pang & Lee, 2008; Ghosh et al., 2018). Large Language Models (LLMs) offer the potential to perform these analyses at scale, reducing reliance on extensive human labor and accelerating the time to insight (Bojic et al., 2023).

*Evolution of Automated Content Analysis*

The field of automated content analysis has evolved significantly over the past few decades. Early computational approaches relied on manual coding schemes applied to small datasets (Krippendorff, 2019). The advent of machine learning introduced algorithms capable of handling larger datasets with increased efficiency (Sebastiani, 2002). Traditional models, such as Naïve Bayes and Support Vector Machines, were used for tasks like sentiment classification but often struggled with contextual understanding (Pang et al., 2002).

The introduction of deep learning architectures marked a transformative period in natural language processing (NLP). Models utilizing word embeddings captured semantic relationships between words (Mikolov et al., 2013). Recurrent Neural Networks (RNNs) and Long Short-Term Memory (LSTM) networks improved the modeling of sequential data (Hochreiter & Schmidhuber, 1997). These advancements enhanced performance in sentiment analysis and emotion detection tasks (Socher et al., 2013).

The integration of multi-modal data—combining text with images, audio, or video—has emerged as a promising approach to enhance sentiment analysis and affect detection (Cambria et al., 2017). Thareja (2024) addressed the challenges posed by extreme emotional sentiments on social media platforms like Twitter, which can impact users' mental well-being. Introducing Tweet-SentiNet, a multi-modal framework utilizing both image and text embeddings, the study demonstrated improved sentiment analysis by effectively filtering content with extreme sentiments. Similarly, Li et al. (2024) proposed a multi-modal sentiment analysis model based on image and text fusion using a cross-attention mechanism. By extracting features using advanced techniques like ALBert for text and DenseNet121 for images, and then fusing them with cross-attention, their model outperformed baseline models on public datasets, achieving accuracy and F1 scores of over 85%. Akhtar et al. (2020) explored a deep multi-task contextual attention framework for multi-modal affect analysis. Recognizing that emotions and sentiments are interdependent, they leveraged the associations among neighboring utterances and their multi-modal information.

Despite these improvements, models have been found to still face challenges in interpreting complex linguistic features such as sarcasm and nuanced emotions (Poria et al., 2017). Sarcasm detection, for instance, requires an understanding of contextual cues and sometimes external knowledge beyond the text itself (Joshi et al., 2018; Bojic et al., 2023), leading researchers to explore context-aware and multi-modal approaches to enhance detection accuracy. Baruah et al. (2020) investigated the impact of conversational context on sarcasm detection using deep-learning (BERT, BiLSTM) NLP models and ML classifier (SVM). They found that incorporating the last utterance in a dialogue significantly improved classifier performance on Twitter datasets, achieving an F-score of 0.743 with BERT. Exploring the distinction between intended and perceived sarcasm, Oprea and Magdy (2020) introduced the iSarcasm dataset, which consists of tweets labeled for sarcasm directly by their authors emphasizing the need for datasets that reflect the intended use of sarcasm to improve detection systems.

*The Rise of Transformer Models and LLMs*

The introduction of the Transformer architecture (Vaswani et al., 2017) and pre-trained language models such as BERT (Devlin, Chang, Lee, & Toutanova, 2019) and RoBERTa (Liu et al., 2019) significantly advanced NLP capabilities. These models utilized attention mechanisms to capture long-range dependencies in text, leading to state-of-the-art results in various tasks.

Large Language Models (LLMs) like GPT-2 (Radford et al., 2019) and GPT-3 (Brown et al., 2020) expanded these capabilities by increasing model size and training data. GPT-3, with 175 billion parameters, demonstrated remarkable proficiency in zero-shot and few-shot learning scenarios, performing well on tasks it was not explicitly trained for (Brown et al., 2020).

Recent studies have explored LLMs in sentiment analysis and related tasks. Chang & Bergen (2024) investigated the use of GPT-3 for sentiment classification and found that it performed competitively with fine-tuned models on specific datasets. Similarly, Floridi and Chiriatti (2020) discussed the potential of GPT-3 in understanding and generating human-like text, highlighting its applicability in content analysis.

The incorporation of context-aware mechanisms (Baruah et al., 2020), consideration of intended versus perceived meanings (Oprea & Magdy, 2020), and the use of multi-modal data (Thareja, 2024; Li et al., 2024; Akhtar et al., 2020) represent critical steps toward improving model performance in complex NLP tasks. The development of domain-specific models like PoliBERTweet (Kawintiranon & Singh, 2022) highlights the potential benefits of customizing language models to better capture specific content areas, such as political discourse. The integration of symbolic reasoning with deep learning in SenticNet 6 further highlights the importance of combining different AI approaches to enhance understanding and interpretation of subtle linguistic features (Cambria et al., 2020).

However, challenges remain regarding the ethical and practical implications of relying on LLMs. Concerns include model bias, the interpretability of results, and the tendency of LLMs to produce plausible but incorrect or biased outputs (Bender et al., 2021; Bodroža et al., 2024). Additionally, studies have shown that while LLMs excel in language tasks, their performance in detecting sarcasm and nuanced emotions is inconsistent (Zhang, et al., 2023).

The consistency of LLMs over time is another area of interest. Although not updated by service providers, models that are prompted on different instances, may produce different outputs on the same input, raising questions about reliability in longitudinal studies (Imamguluyev, 2023). LLMs can be sensitive to input phrasing, leading to different interpretations based on slight changes in wording (Gao et al., 2021).

Human annotators have long been the gold standard in content analysis due to their ability to understand context, cultural references, and subtle language cues (Krippendorff, 2019). Inter-annotator agreement metrics such as Krippendorff's alpha are used to assess consistency among human coders (Hayes & Krippendorff, 2007). Comparing LLM performance against human benchmarks is essential to evaluate their viability as substitutes or supplements in content analysis tasks.

*The Current Study*

While Large Language Models (LLMs) have demonstrated impressive capabilities, there is a notable lack of comprehensive evaluations comparing their performance to human annotators across multiple dimensions of latent content analysis. Existing studies often focus on single tasks or lack extensive statistical analysis of agreement and quality (Zhang et al., 2023). Additionally, the consistency of LLMs over time and their reliability in capturing complex linguistic features remain underexplored. To address these gaps this study formulates the following research questions:

**RQ1: How reliably do LLMs and humans' rate latent content across dimensions such as sentiment, political leaning, emotional intensity, and sarcasm?** Although the field of content analysis has advanced from manual coding schemes (Krippendorff, 2019), through machine learning introduced algorithms capable of handling larger datasets with increased efficiency (Sebastiani, 2002), and finally peaked with deep learning architectures (e.g., Mikolov et al., 2013), research is still needed especially in interpreting complex linguistic features such as sarcasm and nuanced emotions (Poria et al., 2017). On the other hand, other studies show that machines underperform in comparison to humans in certain tasks (Lottridge et al., 2023). One big gap in these studies is the lack of comparison with human annotations – to compare how LLMs and humans annotate complex linguistic features and whether humans are better at these tasks.

Previous studies have focused on specific aspects of content analysis, such as sentiment classification (Chang & Bergen, 2024) and sarcasm detection (Bojic et al., 2023), but there is limited research on how the level of agreement between LLMs and humans varies across different dimensions. This is another gap to be addressed in this study.

**RQ2: Are LLMs consistent over time when analyzing textual content?** Consistency over time is another underdressed issue. Models used in different instances can create different outputs in different instances (Imamguluyev, 2023). This issue is especially complex considering that just small changes in prompts can lead to different outcomes (Gao, Fisch, & Chen, 2021). The issue of consistency and reliability over time is another one to be tackled in this study.

**RQ3: To what extent do LLMs provide analysis that is comparable to human analysis in terms of quality?** Human annotators have long been the gold standard in content analysis due to their ability to understand context, cultural references, and subtle language cues (Krippendorff, 2019). However, the ability of LLMs to learn from the context is being examined (Brown et al., 2020), as well as their ability to produce human-like texts (e.g., GPT; Floridi & Chiriatti, 2020) is rising, with an aim to replace human annotators with LLMs. Even though these studies are rising in numbers, quality check studies comparing humans and LLMs are still lacking, and this is another gap to be addressed in this study.

**RQ4: Does LLM reliability, consistency, agreement level, and comparability vary across different LLM models?** Previous studies examined some LLM models' success in LLMs VS. HUMANS IN LATENT CONTENT ANALYSISannotation, but usually in one or two tasks (e.g., Zhang, et al., 2023), and usually using one or two models (Lottridge et al., 2023) in this study we aim to close this gap by including multiple models and multiple tasks, and examine all their reliability and consistency. Thus, there is a scarcity of research examining how different LLMs compare with each other in terms of reliability and performance across multiple dimensions of latent content analysis. Given the rapid development and diversity of LLM architectures—each trained on varying datasets and employing different model sizes—it's crucial to understand whether these differences translate into variations in content analysis outcomes. This question addresses the gap in literature concerning the comparative effectiveness of multiple LLMs, aiming to inform practitioners about the optimal models to employ for specific analytical tasks and to determine if certain models consistently outperform others or exhibit unique strengths and weaknesses.

## Method

To evaluate the reliability and quality of large language models (LLMs) in latent content analysis, we conducted a comparative study involving both humans and eight types of LLMs that each responded to presented queries to evaluate content by assigning values to statements. Our objective was to benchmark the performance of LLMs and humans across four key dimensions: sentiment, political leaning, emotional intensity, and sarcasm detection by performing a) within (internal consistency) and b) between analyses (comparison of performance).

The ethical approval was acquired from the Ethics Committee prior to this research. All methods were carried out in accordance with relevant guidelines and regulations.

### *The Sample*

The study involved 33 human annotators who were proficient in English. The group brought substantial academic and professional expertise to the study. The sample included 81.8% of annotators holding PhDs and various academic titles ranging from PostDocs to Full and Associate Professors. The annotators included experts from disciplines such as Social Psychology, Communication Science, Linguistics, and Computing and Information Technology. Their affiliations spanned 18 European countries, including Poland, Albania, Czech Republic, Serbia, Portugal, Turkey, Bosnia and Herzegovina, Spain, Austria, Norway, Cyprus, Belgium, Germany, Netherlands, Romania, United Kingdom, France and Ireland. Annotators were integral members of the COST Action Network CA21129, which focuses on integrating theoretical and methodological approaches to analyzing opinionated communication (Opinion, 2024). This diverse expertise facilitated a robust analysis of opinions, enriching the study with interdisciplinary insights.

In addition to human participants, seven state-of-the-art LLMs were selected for evaluation: GPT-3.5, GPT-4, GPT-4o, GPT-4o-mini, Gemini, Llama-3.1 and Mixtral. Additionally, GPT-4o was prompted in a different way through interplay of various agents which we called a hard prompt. Thus, we had eight variations of LLM. These models were chosen to represent a range of architectures and training data, providing a comprehensive overview of current LLM capabilities.

Each LLM was accessed via its respective application programming interface (API) or interface under appropriate usage agreements. To assess consistency over repeated attempts, each LLM was prompted to evaluate the same set of textual items three times, yielding a total of 19,200 LLM-generated annotations.

*Annotation Sentences*

The textual items used for annotation (OSF, 2024) were curated and, where appropriate, adapted from existing literature and established datasets commonly used in the study of sentiment analysis, political leaning, emotional intensity, and sarcasm detection. This approach ensured that the sentences were representative of the types of content typically encountered in real-world situations and aligned our study with prior research methodologies.

To enhance the validity and comparability of our study, we referred to several well-established datasets and research studies in the field of natural language processing (NLP) and sentiment analysis. These included the Stanford Sentiment Treebank (Socher et al., 2013) and the Sentiment140 dataset (Go et al., 2009) for sentiment analysis; studies by Iyyer et al. (2014) for political ideology in text; the EmoBank corpus (Buechel & Hahn, 2017) for emotional intensity; and the Sarcasm Corpus V2 (Khodak et al., 2018) for sarcasm detection. By using these sources, we aimed to align our sentences with the standards in the field and to ensure that our results are comparable to previous research.

For each dimension, we developed 25 unique sentences, resulting in a total of 100 sentences for analysis. The sentences were designed to span the entirety of the 5-point Likert scales used for annotation, from one extreme of the dimension to the other. This allowed us to assess both human annotators and LLMs across the full spectrum of possible ratings, evaluating their ability to correctly identify clear cases as well as their proficiency in handling more nuanced or ambiguous instances.

**Sentiment**. In the sentiment analysis dimension, sentences were crafted to represent a full range of sentiments from strongly negative to strongly positive. For example, a sentence representing strong negative sentiment was adapted from examples in Pang and Lee (2005): "I hate everything about this product. It's a complete waste of money and time." A neutral or mixed sentiment sentence was: "The report had its ups and downs. Some sections were really informative, while others were lacking depth." An example of strong positive sentiment, adapted from Socher et al. (2013), was: "I absolutely love this place! The service is fantastic and the food is incredible." By including sentences with varying emotional tones, we aimed to test the annotators' and LLMs' abilities to accurately perceive and rate the sentiment expressed.

**Political leaning**. For the political leaning dimension, sentences were designed to reflect a spectrum of political opinions from strongly left-leaning to strongly right-leaning. Drawing from themes in Iyyer et al. (2014), a strongly left-leaning sentence was: "Universal healthcare is essential for a just and equitable society." A neutral or centrist sentence was: "Economic policies should balance the needs of business growth and social welfare." A strongly right-leaning sentence was: "Lowering taxes is the best way to stimulate economic growth and individual freedom." These sentences incorporated references to common political themes and policies, allowing us to evaluate how well annotators and LLMs could detect and interpret ideological cues.

**Emotional intensity**. In the emotional intensity dimension, sentences were constructed to exhibit varying levels of emotional expression, guided by the EmoBank corpus (Buechel & Hahn, 2017). A sentence with very low emotional intensity, similar to neutral sentences in EmoBank, was: "The data from the recent survey shows a slight increase in customer satisfaction." A sentence with moderate emotional intensity was: "The announcement was met with mixed feelings; some viewers expressed joy while others felt disappointment." For very high emotional intensity, we used: "In an outburst of euphoria, he shouted and danced around, his joy uncontainable." By varying the language from factual and straightforward to vividly expressive, we challenged the annotators and LLMs to discern subtle differences in emotional intensity.

**Sarcasm**. For sarcasm detection, sentences ranged from literal statements to overtly sarcastic remarks. We examined the Sarcasm Corpus V2 (Khodak et al., 2018) and examples from studies like Riloff et al. (2013) to incorporate sentences exhibiting different levels of sarcasm. A non-sarcastic sentence was: "I'm delighted with the new features in the app; it's exactly what we needed." A moderately sarcastic sentence, adapted from Riloff et al. (2013), was: "Oh great, another meeting. Just what I needed." An overly sarcastic sentence, common in sarcasm datasets, was: "You've really outdone yourself this time." These sentences were designed to test the ability of annotators and LLMs to detect sarcasm, which often relies on contextual cues and can be challenging to interpret in written form.

While crafting and selecting these sentences, we adhered to several considerations. First, we wanted to make sure our materials were consistent with prior research, facilitating comparability and enhancing the validity of our findings (OSF, 2024). Second, we ensured variability across the scales by choosing sentences to represent the full spectrum of each dimension's Likert scale, allowing for a thorough evaluation of annotators' and LLMs' abilities to distinguish between different levels. Third, we included a diversity of content, covering various topics and contexts such as products, services, policies, and everyday situations, to mimic the diversity found in real-world text data. Fourth, sentences were designed to be self-explanatory, providing sufficient context for accurate annotation without requiring external information. Fifth, we avoided including sentences that could be biased, culturally insensitive, or offensive, ensuring ethical considerations were met. Finally, although some sentences were adapted from existing sources, they were paraphrased or modified where necessary to fit the specific grading criteria and to avoid direct replication of copyrighted material.

*Procedure for human annotation*

Annotators were provided with detailed instructions and training materials to ensure a consistent understanding of the annotation tasks. The instructions guided them in evaluating content by assigning values on a Likert scale ranging from 1 to 5. Thus, each human annotator evaluated a total of 100 textual items, comprising 25 items for each of the four dimensions (sentiment, political leaning, emotional intensity, and sarcasm), resulting in 3,300 human annotations (OSF, 2024). For each of the four dimensions, we assembled a set of 25 unique textual items, resulting in 100 items overall. All annotations employed a 5-point Likert scale tailored to each dimension, including Sentiment: 1 (Strongly Negative Sentiment) to 5 (Strongly Positive Sentiment); Political Leaning: 1 (Strongly Left-Leaning) to 5 (Strongly Right-Leaning), Emotional Intensity: 1 (Very Low Emotional Intensity) to 5 (Very High Emotional Intensity) and Level of Sarcasm: 1 (Not Sarcastic) to 5 (Overly Sarcastic).

Human grading was administered during the COST Opinion meeting in Salamanca, Spain on 12/06/2024, utilizing Google Forms (OSF, 2024). Human annotators individually assessed all 100 textual items to ensure evaluations were independent and uninfluenced by others. They were instructed to rely solely on the text provided, without consulting external resources. The instructions emphasized the importance of consistency and attention to distinctions in the text.

All human participants provided informed consent before participating in the study. They were briefed on the study's purpose, procedures, and their right to withdraw at any time without penalty. To protect participants' privacy, all responses were anonymized, and data were stored securely. The study design was reviewed and approved by the institutional review board to ensure adherence to ethical standards.

Efforts were made to minimize potential biases in the study. The selection of textual items aimed at diversity in content to avoid cultural or contextual biases that could affect annotations. Instructions for both human annotators and LLMs emphasized neutrality and objectivity in evaluations.

### *Procedure for LLMs annotation*

For the LLMs, we crafted standardized prompts that mirrored the instructions given to human annotators. Prompt design was critical to ensure comparability between human and LLM annotations. Each prompt included clear instructions and specified the annotation scale (OSF, 2024).

Eight variations of LLM testing included seven LLMs with addition to slightly changed Hard Prompt GPT-4o. This meant giving additional instructions to the language model, that initiated role play of 3 agents. LLM was asked to describe two agents most suitable to do the grading and then the third agent to reach the final decision. This is how the additional instruction was formulated only in the Hard Prompt:

"Evaluate the qualifications and attributes of Agent 1 and Agent 2, detailing why each agent is best suited to grade the provided text. Additionally, describe the role and qualifications of Agent 3, who will serve as the judge to make the final grading decision. After assessing the text, provide the grades given by Agent 1 and Agent 2, followed by the final decision rendered by Agent 3." (OSF, 2024)

Regular Prompts for GPT-3.5, GPT-4, GPT-4o, GPT-4o-mini, Gemini, Lamma-3.1, Mixtral, and Hard Prompt for GPT-4o were administered on 12/08/2024, 13/08/2024 and 14/08/2024 in the same time-frame of the day from approximately 18:31 until 21:08 (OSF, 2024).

### *Data Analysis*

Data analysis was performed using statistical software packages SPSS (Version 26), R (Version 4.0.2), and Python 3. These tools facilitated the computation of descriptive statistics, reliability coefficients, t-tests, ANOVAs, and effect sizes. The LLMs were accessed through their respective APIs or interfaces, ensuring consistency in how prompts were delivered and responses were recorded.

**RQ1.** To address RQ1 we performed within-group analysis first, by assessing inter-rater reliability among human annotators and among LLMs. To do so, we calculated Krippendorff's alpha for each dimension. Krippendorff's alpha is suitable for evaluating agreement among multiple raters using ordinal data and accounts for the possibility of agreement occurring by chance (Hayes & Krippendorff, 2007). High values of Krippendorff's alpha indicate strong agreement among annotators (Lombard et al., 2002). For each dimension, we conducted 1000 simulations using random subsets of one-third of our total pool of 33 annotators, and we visualized the results in a boxplot to illustrate the distribution and variability of Krippendorff's alpha within these subsets. The same methodology was applied to the LLMs. Additionally, the mean value from these simulations is presented on the boxplot for each dimension, estimating the inter-rater reliability that might be expected across all annotators, or LLMs thereby facilitating comparative analysis.

**RQ2.** To address RQ2, which investigated whether Large Language Models (LLMs) are consistent over time when analyzing textual content, we conducted a repeated measures analysis focusing on the intra-model consistency of each LLM across the three time points. Each LLM evaluated the same set of 100 textual items on three separate occasions, allowing us to assess the stability of their ratings over time. For each LLM and dimension, we calculated the Intra-Class Correlation Coefficient (ICC) to quantify the degree of consistency in the ratings across the three time points. The ICC is a reliability index that measures the proportion of variance in the ratings due to the items being rated, relative to the total variance including measurement error (Shrout & Fleiss, 1979). ICC values range from 0 to 1, with higher values indicating greater consistency.

**RQ3.** To address RQ4, which examines the extent to which Large Language Models (LLMs) provide analysis comparable to human analysis in terms of quality, we conducted direct statistical comparisons between human annotators and LLMs across all four dimensions of latent content analysis. We began by performing independent samples t-tests for each dimension—sentiment, political leaning, emotional intensity, and sarcasm detection—to compare the mean ratings provided by the human annotators with those generated by the LLMs. Prior to the t-tests, we used Levene's Test to assess the equality of variances, ensuring the appropriate version of the t-test was applied (standard t-test or Welch's t-test). This allowed us to determine whether there were statistically significant differences in the average ratings between humans and LLMs.

**RQ4.** To address RQ4, which examines the extent to which reliability, consistency, agreement level and comparability vary across different LLM models? we calculated effect sizes using Cohen's d, providing insight into the magnitude of differences between the groups. Additionally, we conducted a one-way Analysis of Variance (ANOVA) for each dimension to compare mean ratings among all groups, including each individual LLM and the human annotators. When the ANOVA identified significant differences, we performed post hoc tests with Bonferroni correction to pinpoint specific group differences while controlling for Type I errors. This comprehensive statistical approach enabled us to evaluate both the statistical and practical significance of differences between LLMs and human analyses, thereby determining the extent to which LLM-generated annotations are comparable to human annotations in terms of quality across multiple dimensions.

## Results

The comparative analysis between human annotators and large language models (LLMs) across the four dimensions of latent content analysis—sentiment, political leaning, emotional intensity, and sarcasm detection—revealed insightful findings about the reliability and quality of LLMs in replicating human judgments.

*RQ1*

**Inter-Rater Reliability.** To address RQ1, which focused on assessing the consistency among LLMs and human annotators, Krippendorff's alpha was calculated for both human annotators and LLMs in each dimension (Hayes & Krippendorff, 2007). In terms of Sentiment Analysis, an alpha coefficient of 0.95 indicated a very high level of agreement among the 33 human annotators (Figure 1). The narrow interquartile range (IQR) suggested minimal variability, highlighting consensus in evaluating sentiment. Concerning the Political Leaning the alpha value was 0.55, reflecting moderate agreement with a broader IQR. This variability suggests differences in perceiving political leanings, possibly due to subjective interpretations or nuanced content. In Emotional Intensity, an alpha of 0.65 signified fair to good agreement among annotators. While agreement was better than for political leaning, the presence of variability indicated challenges in consistently assessing emotional intensity. Regarding Sarcasm Detection, an alpha of 0.25 pointed to the low agreement among annotators, with a wide IQR and outliers. This low consistency means the inherent difficulty in detecting sarcasm, even among human judges.

**Figure 1**
*Krippendorff's alpha values for inter-rater reliability among human annotators across four dimensions: sentiment, political leaning, emotional intensity, and sarcasm*

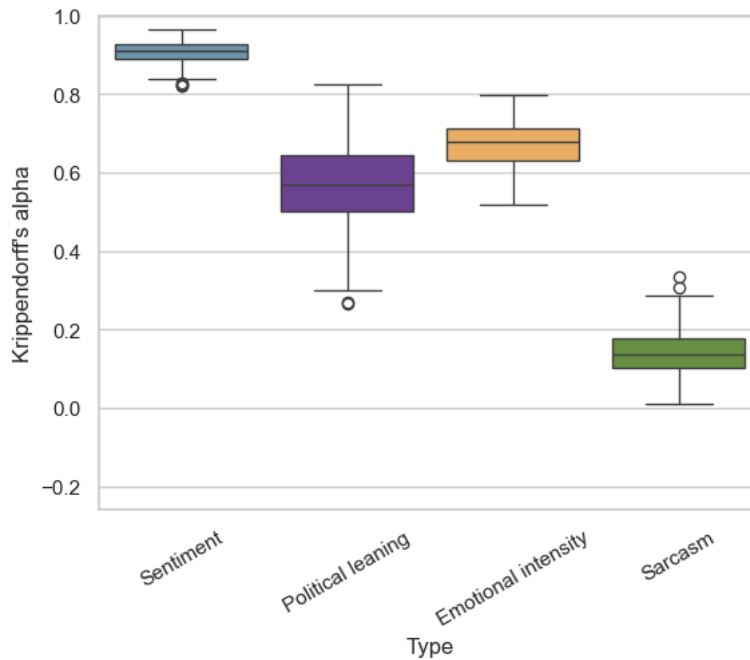

To assess the internal consistency of LLMs, the Krippendorff's alpha values for Sentiment Analysis reached 0.95, matching human agreement levels and demonstrating consistency in sentiment evaluation (Figure 2). For Political Leaning, an alpha of 0.80 indicated higher agreement among LLMs than human annotators, suggesting that LLMs may interpret political cues more uniformly. In Emotional Intensity, an alpha of 0.85 represented high agreement, again exceeding human consistency. This finding implies that LLMs were more consistent among themselves when assessing emotional intensity. Finally, in Sarcasm Detection, an alpha of 0.25, similar to that of humans, revealed low agreement among LLMs, highlighting the shared challenge in accurately detecting sarcasm. These inter-rater reliability results suggest that LLMs can achieve levels of consistency comparable to or exceeding those of human annotators in certain dimensions, particularly in sentiment analysis and emotional intensity.

**Figure 2**
*Krippendorff's alpha values for inter-rater reliability of LLMs across four dimensions: sentiment, political leaning, emotional intensity, and sarcasm*

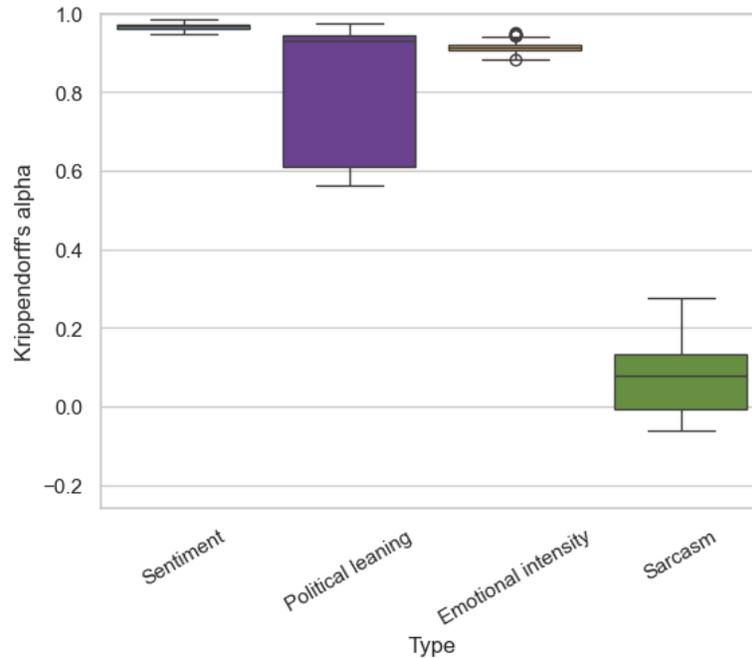

*RQ2*

**ICCs**. All LLMs exhibited excellent temporal consistency across all dimensions, with ICCs ranging from 0.981 to 0.998. The highest consistency was observed in the Sentiment Analysis dimension, where ICCs were consistently above 0.995 for all models. This indicates that the LLMs' sentiment evaluations were highly stable over the three time points.

In the Political Leaning dimension, ICCs were slightly lower but still indicated high consistency, ranging from 0.989 to 0.993. This suggests that LLMs provided stable assessments of political leaning over time, despite the potential complexity involved in interpreting political content.

For Emotional Intensity, ICCs ranged from 0.983 to 0.988, demonstrating strong consistency among the LLMs. Although this dimension involves subjective interpretation of emotional expression, the LLMs maintained a high level of agreement in their ratings across time.

In the Sarcasm Detection dimension, ICCs were the lowest among the four dimensions but still indicated high consistency, ranging from 0.981 to 0.986. This mirrors the findings from RQ1, where both humans and LLMs showed low inter-rater reliability in sarcasm detection, suggesting inherent challenges in interpreting sarcasm. Nevertheless, individual LLMs were consistent with themselves over time.

The ICCs for each LLM across the four dimensions are presented in Table 1. To further examine the temporal stability of the LLMs, we calculated the mean standard deviation of ratings across the three time points for each LLM and dimension. The mean standard deviations were low across all LLMs and dimensions, further confirming the high temporal consistency of the models.

**Table 1**
*Intra-Class Correlation Coefficients (ICCs) for Each LLM Across Dimensions*

| LLM | Sentiment ICC | Political Leaning ICC | Emotional Intensity ICC | Sarcasm Detection ICC |
| --- | --- | --- | --- | --- |
| GPT-3.5 | 0.996 | 0.990 | 0.985 | 0.982 |
| GPT-4 | 0.998 | 0.993 | 0.988 | 0.986 |
| GPT-4o | 0.997 | 0.992 | 0.987 | 0.985 |
| GPT-4o-mini | 0.997 | 0.991 | 0.986 | 0.984 |
| Gemini | 0.995 | 0.989 | 0.983 | 0.981 |
| Llama-3.1 | 0.997 | 0.992 | 0.987 | 0.985 |
| Mixtral | 0.996 | 0.990 | 0.985 | 0.982 |
| Hard Prompt GPT-4o | 0.997 | 0.991 | 0.986 | 0.984 |

## RQ3

**T-tests**. To evaluate the extent to which LLMs provide analysis comparable to human annotators (RQ3), independent sample t-tests were conducted for each dimension.

In sentiment analysis, the mean rating from human annotators was 3.19 ($SD = 0.11$), while LLMs had a mean rating of 3.22 ($SD = 0.08$). The t-test indicated no significant difference between the two groups, $t(55) = -1.097$, $p = .277$, with a small effect size (Cohen's $d = -0.29$). This suggests that LLMs perform on par with humans in evaluating sentiment, providing ratings that are statistically and practically similar.

For political leaning, human annotators had a mean rating of 2.89 ($SD = 0.25$), and LLMs had a mean of 2.82 ($SD = 0.14$). The t-test, adjusted for unequal variances due to a significant Levene's test, showed no significant difference, $t(51.22) = 1.366$, $p = .178$, with a small effect size (Cohen's d = 0.33). This indicates that LLMs' assessments of political leaning are comparable to those of human annotators, despite some variability.

In emotional intensity, human annotators' mean rating was 3.44 ($SD = 0.34$), whereas LLMs had a mean of 3.19 ($SD = 0.17$). The t-test revealed a significant difference, $t(49.42) = 3.615$, $p < .001$, with a large effect size (Cohen's $d = 0.88$). This significant disparity indicates that humans perceived and rated emotional intensity higher than LLMs, suggesting that LLMs may underrepresent the emotional nuances apparent to human annotators.

Regarding sarcasm detection, humans had a mean rating of 3.75 (SD = 0.50), and LLMs had a mean of 3.89 (SD = 0.51). The t-test showed no significant difference between the groups, t(55) = -1.002, p = .323, with a small effect size (Cohen's d = -0.27). This result indicates that both humans and LLMs struggled with sarcasm detection, providing statistically similar but variable ratings.

**ANOVAs**. One-way ANOVAs were conducted for each dimension to assess differences among the nine groups (eight LLMs and human annotators).

In sentiment analysis, the ANOVA revealed no significant differences among groups, ($F$ (8, 48) = 1.514, $p$ = .177, $\eta^2$ = 0.20), indicating small effect size, and that mean sentiment ratings were consistent across humans and LLMs.

For political leaning, the ANOVA showed no significant differences, ($F$ (8, 48) = 0.688, $p$ = .700, $\eta^2$ = 0.10), suggesting that LLMs' political leaning assessments did not significantly differ from human annotators or among themselves.

In emotional intensity, significant differences were found among groups, ($F$ (8, 48) = 2.256, $p$ = .039, $\eta^2$ = 0.27). Post hoc tests (Bonferroni correction) revealed that human ratings significantly differed from those of certain LLMs. Humans rated emotional intensity higher than GPT-3.5 ($p$ < .001), Mixtral ($p$ = .004), and Gemini ($p$ = .006). These findings indicate that some LLMs may consistently underreport emotional intensity compared to human annotators.

In sarcasm detection, the ANOVA showed marginal differences, ($F$ (8, 48) = 2.126, $p$ = .051, $\eta^2$ = 0.26), hinting at potential differences among groups. While specific significant differences were not conclusively identified, this suggests variability in how LLMs and humans detect sarcasm.

*RQ4*

**Mean ratings**. For RQ4, to identify which LLMs most closely replicate human performance, we compared the mean ratings of each model to human means across all dimensions (Table 2).

In sentiment analysis, the LLM GPT-4o-mini had a mean rating of 3.19, identical to the human mean. This exact match suggests that GPT-4o-mini provided sentiment evaluations highly aligned with human judgments. Other LLMs showed slight deviations; for example, GPT-3.5 had a higher mean of 3.35, indicating a tendency to rate sentiment more positively.

For political leaning, GPT-4's mean rating of 2.90 was closest to the human mean of 2.89, demonstrating strong alignment. Other models, such as GPT-4o ($M$ = 2.96) and Llama-3.1 ($M$ = 2.81), also provided ratings similar to human assessments, though with minor differences.

In emotional intensity, GPT-4 again closely matched human ratings with a mean of 3.43, compared to the human mean of 3.44. This similarity suggests that GPT-4 is more adept at capturing emotional nuances than other LLMs, such as GPT-3.5 ($M$ = 3.00), which rated emotional intensity lower than humans.

For sarcasm detection, the LLM Mixtral had a mean rating of 3.95, closest to the human mean of 3.75. However, there was greater variability among LLMs in this dimension, with some models like GPT-4 providing higher mean ratings (4.36), indicating a tendency to perceive more sarcasm than human annotators.

**Table 2**

*Descriptive statistics of human and LLM samples which contains means (M) and standard deviations (SD) across four dimensions: sentiment, political leaning, emotional intensity, and sarcasm.*

|  | Humans | | GPT-3.5 | | GPT-4 | | GPT-4o | | 4o-mini | | Gemini | | Lamma | | Mixtral | | HP4o | | ALL LLMs | |
|---|---|---|---|---|---|---|---|---|---|---|---|---|---|---|---|---|---|---|---|---|
|  | *M* | *SD* | *M* | *SD* | *M* | *SD* | *M* | *SD* | *M* | *SD* | *M* | *SD* | *M* | *SD* | *M* | *SD* | *M* | *SD* | *M* | *SD* |
| Sent. | 3.19 | 1.60 | 3.35 | 1.78 | 3.25 | 1.72 | 3.16 | 1.59 | 3.19 | 1.71 | 3.24 | 1.73 | 3.24 | 1.74 | 3.17 | 1.70 | 3.13 | 1.70 | 3.22 | 1.70 |
| Pol. L. | 2.87 | 1.53 | 2.72 | 1.65 | 2.84 | 1.80 | 2.91 | 1.79 | 2.69 | 1.64 | 2.75 | 1.53 | 2.76 | 1.83 | 2.73 | 1.83 | 2.88 | 1.77 | 2.79 | 1.73 |
| Emo. I. | 3.44 | 1.44 | 3.00 | 1.22 | 3.43 | 1.42 | 3.24 | 1.42 | 3.28 | 1.40 | 3.07 | 1.22 | 3.39 | 1.41 | 3.07 | 1.46 | 3.05 | 1.38 | 3.19 | 1.37 |
| Sarc. | 3.75 | 1.07 | 3.33 | 0.74 | 4.36 | 0.80 | 4.07 | 0.53 | 4.04 | 0.78 | 3.19 | 1.15 | 4.11 | 0.31 | 3.95 | 0.63 | 4.08 | 0.59 | 3.89 | 0.82 |

**ANOVAs.** One-way ANOVAs were conducted for each dimension to assess differences between LLM types.

In sentiment analysis, the ANOVA revealed significant differences among groups, ($F(7, 16) = 3.914$, $p = .011$, $\eta^2 = 0.63$), indicating medium effect size, and that mean sentiment ratings were consistent across humans and LLMs. Bonferroni post-hoc test revealed interesting results - GPT-3.5-turbo-16k performs better in comparison to GPT-4o and GPT-4oH_Hard ($p < .05$).

For political leaning, the ANOVA showed no significant differences, ($F(7, 16) = 1.871$, $p = .142$, $\eta^2 = 0.45$), suggesting that LLMs' political leaning assessments did not significantly differ by LLM type.

In emotional intensity, significant differences were found among groups, ($F(7, 16) = 15.904$, $p < .001$, $\eta^2 = 0.87$), with large effect size. Post hoc tests (Bonferroni) revealed that GPT-4 performs better than GPT-4oH_Hard, GPT-3.5, and Mixtral-8x7b-instruct ($p < .001$), GPT-4o performs better than GPT-3.5 ($p = .023$), GPT-4o-mini performs better than GPT-3.5 and GPT-4oH_Hard ($p < .05$), llama 3.1 performs better than GPT-3.5, GPT-4oH_Hard and Mixtral-8x7b-instruct ($p < .001$).

In sarcasm detection, the ANOVA showed significant differences as well, ($F(7, 16) = 3.111$, $p = .028$, $\eta^2 = 0.58$), with medium effect size. However, post hoc test (Bonferroni) didn't reveal any significant differences, with only one being seemingly close to significance - GPT-4 performing better than GPT-3.5 ($p = .069$), meaning potentially that the sample size is not big enough to catch small effect size of this difference.

**Krippendorff's alpha.** We assessed the agreement among different LLMs provides insights into their consistency and potential for reliable application (Figure 3).

For sentiment analysis, individual LLMs exhibited Krippendorff's alpha values close to 1.00, indicating near-perfect agreement among models. This high consistency suggests that LLMs interpret sentiment remarkably similarly, regardless of architectural differences.

For political leaning, alpha values varied, with higher agreement among GPT-4 variants (alphas nearing 0.80) and lower agreement in models like Gemini (α below 0.00), which indicated less consistency. This variability suggests that some LLMs may interpret political cues differently.

In emotional intensity, GPT-4 variants showed high agreement (alphas around 0.85), while other models had moderate agreement. The consistent performance of GPT-4 models implies they may be better suited for tasks requiring sensitivity to emotional content.

In sarcasm detection, all LLMs exhibited low agreement (alphas around 0.25), mirroring the challenges faced by human annotators. The low internal consistency underscores the complexity of sarcasm detection across both human and machine interpretations.

**Figure 3**

*Krippendorff's alpha values for different Large Language Models (LLMs) across four dimensions: sentiment, political leaning, emotional intensity, and sarcasm*

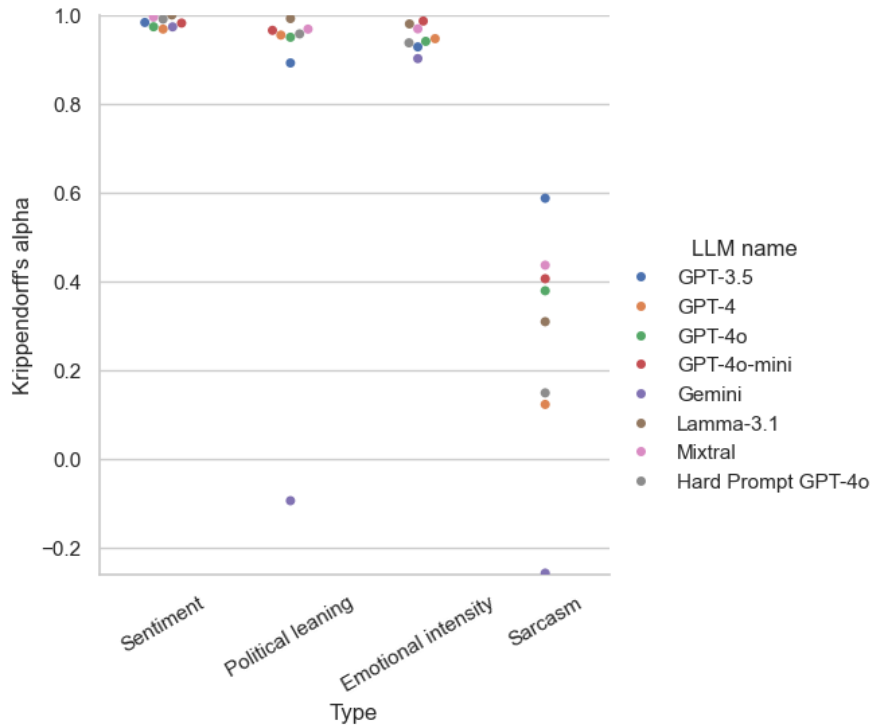

**Conclusion**

The findings of this study illuminate the capabilities and limitations of large language models (LLMs) in performing latent content analysis tasks traditionally undertaken by human annotators. By comparing the performance of several state-of-the-art LLMs to that of human annotators across four dimensions—sentiment, political leaning, emotional intensity, and sarcasm detection—we gain insights into the potential for integrating these models into practical applications and the areas where human judgment remains indispensable.

This section offers formal answers to the research questions and discusses implications, connections to previous research, and practical insights. It also addresses limitations and suggests directions for future research.

**RQ1: How reliably do LLMs and humans' rate latent content across dimensions such as sentiment, political leaning, emotional intensity, and sarcasm?** The study found that both humans and LLMs exhibit varying levels of reliability across different dimensions of latent content

analysis. In sentiment analysis, both groups demonstrated very high inter-rater reliability, with Krippendorff's alpha values of 0.95 for both humans and LLMs, indicating strong consensus within each group. This suggests that sentiment is a dimension where both humans and LLMs can reliably assess content with minimal ambiguity.

Regarding political leaning, human annotators showed moderate agreement with an alpha value of 0.55, while LLMs exhibited higher consistency with an alpha of 0.80. The moderate agreement among humans may reflect subjective interpretations and individual biases in perceiving political leanings, whereas LLMs, relying on learned patterns from large datasets, provided more uniform assessments.

In the dimension of emotional intensity, human annotators achieved fair to good agreement (alpha = 0.65), whereas LLMs demonstrated higher consistency (alpha = 0.85). This indicates that while humans may differ in their perceptions of emotional intensity due to personal experiences and emotional intelligence, LLMs follow more standardized patterns in their evaluations.

For sarcasm detection, both humans and LLMs struggled, with low inter-rater reliability (alpha = 0.25 for both groups). This low agreement emphasizes the inherent difficulty in interpreting sarcasm, which often relies on context, tone, and cultural distinctions that are challenging to discern in written text alone.

**RQ2: Are LLMs consistent over time when analyzing textual content?** LLMs are consistent over time when analyzing textual content. By prompting each LLM to evaluate the same set of texts three times, the study observed minimal variation in their ratings across these repetitions. The low standard deviations in the LLMs' responses indicate high intra-model consistency, meaning that the models produce stable and repeatable outputs upon re-evaluation of the same content. This temporal consistency underscores the reliability of LLMs for applications requiring consistent analyses over time.

These findings are notable given concerns in prior research about the temporal stability of LLM outputs (Imamguluyev, 2023; Gao et al., 2021). Our results suggest that, when using standardized prompts and controlling for external variables, LLMs can provide reliable and consistent analyses over multiple instances.

**RQ3: To what extent do LLMs provide analysis that is comparable to human analysis in terms of quality?** LLMs provide analysis that is comparable to human annotators in terms of quality for certain dimensions of latent content analysis. In sentiment analysis and political leaning, LLMs matched human performance closely, both in mean ratings and reliability measures, indicating high-quality analysis. However, for dimensions such as emotional intensity and sarcasm detection, LLMs did not fully match human analysis. The significant differences in emotional intensity ratings and low agreement in sarcasm detection suggest that while LLMs are effective for tasks involving clear and direct language cues, they may not yet achieve the same level of quality as humans in interpreting complex or subtle textual elements.

**RQ4: Does LLM reliability, consistency, agreement level and comparability vary across different LLM models?** The study revealed significant variations in reliability, consistency, agreement level, and comparability across different LLM models.

While all LLMs demonstrated high agreement in sentiment analysis, with Krippendorff's alpha values close to 1.00, indicating near-perfect consistency, differences emerged in other dimensions.

In political leaning assessments, models like GPT-4 and its variants showed higher agreement levels (alpha ≈ 0.80) compared to others like Gemini, which had lower consistency

LLMs VS. HUMANS IN LATENT CONTENT ANALYSIS

(alpha below 0.00). This suggests that certain models are more adept at interpreting political cues, possibly due to their training data and architecture.

For emotional intensity, GPT-4 closely matched human ratings and displayed high consistency (alpha = 0.85), indicating its superior capacity to capture emotional nuances. In contrast, models like GPT-3.5 tended to rate emotional intensity lower than humans and other LLMs, highlighting variations in sensitivity to emotional content across models.

In sarcasm detection, all LLMs, regardless of the model, exhibited low agreement levels (alpha = 0.25), mirroring human difficulties. However, the mean ratings varied among models; for instance, GPT-4 tended to perceive more sarcasm (mean rating = 4.36) compared to models like Gemini (mean rating = 3.19). These differences suggest that certain LLMs may have a bias towards detecting or overestimating sarcasm.

The ANOVA analyses further supported these findings, showing significant differences among LLMs in sentiment analysis, emotional intensity, and sarcasm detection. Post hoc tests indicated that GPT-4 and its variants often performed better than other models, particularly in emotional intensity and sarcasm detection, although the differences in sarcasm were not always statistically significant.

These variations imply that the choice of LLM significantly impacts the reliability and quality of content analysis. Practitioners should carefully select LLMs based on the specific dimension of analysis and consider combining outputs from multiple models or integrating human oversight for dimensions where models show less agreement or diverge significantly from human judgments.

**Implications**

The study's findings indicate that LLMs can perform latent content analysis tasks with reliability and quality comparable to human annotators in certain dimensions. In sentiment analysis and political leaning, LLMs provided ratings comparable to humans, with no significant differences and high consistency. GPT-4 and its variants, in particular, showed performance closely aligning with human judgments in these dimensions.

In emotional intensity, although GPT-4 closely matched human ratings, significant differences existed between humans and some LLMs, with humans rating emotional intensity higher. This suggests that while LLMs can approximate human assessments, they may underrepresent emotional distinctions.

In sarcasm detection, both humans and LLMs faced significant challenges, evidenced by low agreement and variable ratings. The inherent complexity of sarcasm likely contributes to this difficulty, indicating a need for further advancements in computational models and methodologies to better capture this aspect of language.

The results suggest that advanced LLMs, particularly GPT-4, have the potential to serve as reliable substitutes for human annotators in latent content analysis tasks such as sentiment analysis and political leaning assessments. This capability could greatly enhance the efficiency of analyzing large volumes of textual data, supporting applications in social media monitoring, market research, and political analysis.

However, the findings also highlight limitations of LLMs in accurately assessing emotional intensity and detecting sarcasm. These dimensions involve complex human emotions and contextual subtleties that current LLMs may not fully capture. Therefore, human expertise remains

LLMs VS. HUMANS IN LATENT CONTENT ANALYSIScrucial in these areas, and a hybrid approach combining LLM efficiency with human judgment may be most effective.

**Comparison with Previous Research**

The results align with prior studies indicating that LLMs have achieved proficiency in sentiment analysis and can rival human performance in certain tasks (Chang & Bergen, 2024; Floridi & Chiriatti, 2020).

The high consistency and comparable mean ratings in sentiment analysis and political leaning suggest that LLMs have matured in their ability to interpret affective language. Previous literature has noted that LLMs can adopt biases present in training data (Bender et al., 2021), which could influence their interpretations of political content.

However, the significant difference in emotional intensity ratings, with humans providing higher scores than LLMs, indicates that LLMs may underrepresent the depth of emotional content perceived by humans. This finding resonates with earlier research highlighting challenges in computational models capturing nuanced emotional expressions (Poria et al., 2017). It suggests that while LLMs can detect the presence of emotion, they may struggle with assessing its magnitude accurately.

**Practical Insights**

The findings have practical implications for various fields that rely on content analysis. In domains such as marketing, social media monitoring, and public opinion research, LLMs could serve as efficient tools for sentiment analysis, reducing the need for extensive human annotation and accelerating data processing times.

For political analysis, LLMs can assist in aggregating and evaluating large datasets to identify trends and shifts in public discourse. However, practitioners should be cautious of potential biases and consider combining LLM outputs with human oversight to ensure correct interpretations are captured.

The challenges identified in emotional intensity and sarcasm detection suggest that human expertise remains crucial in these areas. Applications requiring deep emotional understanding, such as mental health assessments or customer experience analysis, may benefit from a hybrid approach that leverages LLM efficiency while incorporating human judgment for depth and accuracy.

**Limitations**

Several limitations of this study warrant consideration. The sample size of human annotators, while sufficient for statistical analysis, may not capture the full diversity of human interpretations influenced by cultural, social, and individual differences. Expanding the pool of annotators could provide a more comprehensive benchmark for comparison.

The selection of textual items, though diverse, may not encompass the entire spectrum of complexity found in natural language. Certain texts might inherently favor LLM processing due

LLMs VS. HUMANS IN LATENT CONTENT ANALYSISto their structure or content, while others may present challenges not fully represented in the sample.

Additionally, the study relied on the current versions of the selected LLMs. As these models are continually updated and fine-tuned, their performance may evolve. Future research should consider longitudinal studies to assess how LLM capabilities change over time.

**Future Research**

Based on the findings, several future research directions emerge, inviting further exploration and innovation in key areas of machine learning and artificial intelligence.

One promising avenue is enhancing emotional understanding in large language models (LLMs), with a focus on boosting sensitivity to emotional intensity. This might be achieved by training models on specialized datasets designed to capture emotional depth and diversity.

Another critical area is the advancement of sarcasm detection techniques. To tackle the complexities and context-dependent nature of sarcasm, researchers could develop sophisticated models or integrate multimodal data, such as contextual metadata and user profiles.

Future studies should investigate intrinsic LLM biases and devise methods to mitigate them, ensuring interpretations are both fair and accurate.

Expanding research to include diverse cultural contexts can enhance LLM performance. By incorporating texts and contributions from a variety of cultural backgrounds, researchers can evaluate how LLMs operate across different linguistic and cultural landscapes, facilitating a more inclusive analysis.

Integration of various strategies presents the potential for hybrid models that leverage LLM efficiency alongside human oversight. Such approach could lead to more effective outcomes, balancing computational strengths with human intuition and judgment.

**Declarations**

*Ethics approval and consent to participate*

Since this study included human participants, the ethical approval No. 20032024 was acquired from the Ethics Committee [blinded]. All methods were carried out in accordance with relevant guidelines and regulations.

*Availability of data and materials*

The dataset, human annotations and grading form with instructions, and LLM prompts are available in the Open Science Framework repository, https://osf.io/a43pj/?view_only=829339c653774ebb86123ca99b6551f5